\documentclass[11pt,a4paper]{article}
\usepackage[hyperref]{emnlp-ijcnlp-2019}
\usepackage{times}
\usepackage{array,multirow,graphicx}
\usepackage{url}
\usepackage{comment}
\usepackage{enumitem}
\usepackage{tikz}
\usepackage{booktabs}
\usepackage{latexsym}
\usepackage{tabularx}
\usepackage{todonotes}
\usepackage{xspace}
\usepackage{url}
\usepackage{boldline}

\DeclareMathAlphabet{\mathbcal}{OMS}{cmsy}{b}{n}

\aclfinalcopy

\newcommand{\rouge}{\textsc{ROUGE}\xspace}
\newcommand{\rougebest}{\textsc{best-rouge}\xspace}

\newcommand{\bleu}{\textsc{BLEU}\xspace}
\newcommand{\duc}{\textsc{DUC}\xspace}
\newcommand{\ducfour}{\textsc{DUC-04}\xspace}
\newcommand{\ducfive}{\textsc{DUC-05}\xspace}
\newcommand{\ducsix}{\textsc{DUC-06}\xspace}
\newcommand{\ducseven}{\textsc{DUC-07}\xspace}

\newcommand{\bert}{\textsc{BERT}\xspace}

\newcommand{\bertfrlm}{\textsc{bert-fr-lm}\xspace}
\newcommand{\bertfrns}{\textsc{bert-fr-ns}\xspace}

\newcommand{\bertftsone}{\textsc{bert-ft-s-1}\xspace}
\newcommand{\bertftmone}{\textsc{bert-ft-m-1}\xspace}
\newcommand{\bertftmfive}{\textsc{bert-ft-m-5}\xspace}
\newcommand{\bigruatt}{\textsc{B}i\textsc{GRU-ATT}\xspace}
\newcommand{\bigruattsone}{\textsc{b}i\textsc{gru-att-s-1}\xspace}
\newcommand{\bigruattmone}{\textsc{b}i\textsc{gru-att-m-1}\xspace}
\newcommand{\bigruattmfive}{\textsc{b}i\textsc{gru-att-m-5}\xspace}
\newcommand{\bigru}{\textsc{B}i\textsc{GRU}\xspace}
\newcommand{\lm}{\textsc{LM}\xspace}
\newcommand{\gpt}{\textsc{gpt-2}\xspace}
\newcommand{\sumqe}{\textsc{Sum-QE}\xspace}

\title{\sumqe: a BERT-based Summary Quality Estimation Model}

\author{\parbox{12cm}{\centering Stratos Xenouleas$^1$, Prodromos Malakasiotis$^1$, \\
\textbf{Marianna Apidianaki$^2$ \and Ion Androutsopoulos$^1$}} \\
$^1$ Department of Informatics, Athens University of Economics and Business, Greece\\
$^2$ CNRS, LLF, France and University of Helsinki, Finland\\
  {\tt stratosxen@gmail.com, rulller@aueb.gr}\\
  {\tt marianna.apidianaki@helsinki.fi, ion@aueb.gr}
}

\date{}

\begin{document}
\maketitle
\begin{abstract}
We propose \sumqe, a novel Quality Estimation model for summarization based on BERT. The model addresses linguistic quality aspects that are only indirectly captured by  content-based approaches to summary evaluation, without involving comparison with human references. \sumqe achieves very high correlations with human ratings, outperforming simpler models addressing these linguistic aspects. Predictions of the \sumqe model can be used for system development, and to inform  users of the quality of automatically produced summaries and other types of generated text.
\end{abstract}

\section{Introduction}

Quality Estimation (QE) is a term 
used in machine translation (MT) to refer to methods that measure the quality of automatically translated text without relying on human references  \cite{bojar-etal-2016-findings,bojar-etal-2017-findings}. In this study, we address QE for summarization. Our proposed model, \sumqe, successfully predicts linguistic qualities of summaries that traditional evaluation metrics fail to capture \cite{Lin:2004, Lin_Hovy:2003,Papineni:2002,Nenkova:2004}. \sumqe predictions can be used for system development, to inform users of the quality of automatically produced summaries and other types of generated text, and to select the best among summaries output by multiple systems. 

\sumqe relies on the BERT language representation model \cite{devlin-etal-2019-bert}. We use a pre-trained BERT model adding just a task-specific layer, and fine-tune the entire model on the task of predicting linguistic quality scores manually  assigned to summaries. The five criteria addressed are given in Figure~\ref{lqs}. We provide a thorough evaluation on three publicly available summarization datasets from NIST shared tasks, and compare the performance of our model to a wide variety of baseline methods capturing different aspects of linguistic quality. \sumqe achieves very high correlations with human ratings, showing the ability of BERT to model linguistic qualities that relate to both text content and form.\footnote{Our code is available at \url{https://github.com/nlpaueb/SumQE}}

\begin{centering}
\begin{figure}[t!]
\tikz\node[draw=black,thin,inner sep=1pt]{
\scalebox{0.95}{
\footnotesize
    \begin{tabular}{m{0.95\columnwidth}}
    \vspace{1mm}
    {\bf$\mathbcal{Q}$1 -- Grammaticality:} The summary should have no datelines, system-internal formatting,
  capitalization errors or obviously ungrammatical sentences (e.g.,
  fragments, missing components) that make the text difficult to read. \vspace{1mm} \\ 
    {\bf$\mathbcal{Q}$2 -- Non redundancy:} There should be no unnecessary repetition in the summary.\vspace{1mm} \\
    {\bf$\mathbcal{Q}$3 -- Referential Clarity:}  It should be easy to identify who or what the pronouns and noun
      phrases in the summary are referring to.\vspace{1mm} \\
    {\bf$\mathbcal{Q}$4 -- Focus:} The summary should have a focus; sentences should only contain information that is related to the rest of the summary.\vspace{1mm} \\ 
    {\bf$\mathbcal{Q}$5 -- Structure \& Coherence:} The summary should be well-structured and well-organized. The summary should not just be a heap of related information, but should build from sentence to sentence to a coherent body of information about a topic.\\
    \end{tabular}}
};
\caption{\sumqe rates summaries with respect to five linguistic qualities \cite{DUC:2005}. The datasets we use for tuning and evaluation contain human assigned scores (from 1 to 5) for each of these categories.}
\label{lqs}
\end{figure}
\end{centering}

\section{Related Work}

Summarization evaluation metrics like Pyramid \cite{Nenkova:2004} and \rouge \cite{Lin_Hovy:2003,Lin:2004} are recall-oriented; they basically measure the content from a model (reference) summary that is preserved in peer (system generated) summaries. Pyramid requires substantial human effort, even in its more recent versions that involve the use of word embeddings \citep{Passonneau2013} and a lightweight crowdsourcing scheme \citep{Shapira:2019}. \rouge is the most commonly used evaluation metric \cite{Nenkova2012,Allahyari:2017,Gambhir2017}. Inspired by \bleu \cite{Papineni:2002}, it relies on common $n$-grams or subsequences between peer and model summaries. Many \rouge versions are available, but it remains hard to decide which one to use \citep{Graham:2015}. Being recall-based, \rouge correlates well with Pyramid but poorly with linguistic qualities of summaries. \citet{Louis2013} proposed a regression model for measuring summary quality without references. The scores of their model correlate well with Pyramid and Responsiveness, but text quality is only addressed indirectly.\footnote{In the Responsiveness annotation instructions, annotators were asked to assess the linguistic quality of the summary only if it interfered with the expression of information and reduced the amount of conveyed information. See \url{https://duc.nist.gov/duc2005/responsiveness.assessment.instructions}}

Quality Estimation is well established in MT  \cite{callison-burch-etal-2012-findings,bojar-etal-2016-findings,bojar-etal-2017-findings,martins-etal-2017-pushing,W18-6451}. QE methods provide a quality indicator for translation output at run-time without relying on human references, typically needed by MT evaluation metrics \cite{Papineni:2002,denkowski:lavie:meteor-wmt:2014}. QE models for MT make use of large post-edited datasets, and apply machine learning methods to predict post-editing effort scores and quality (good/bad) labels.

We apply QE to summarization, focusing on linguistic qualities that reflect the readability and fluency of the generated texts. Since no post-edited datasets -- like the ones used in MT -- are available for summarization, we use instead the ratings assigned by human annotators with respect to a set of linguistic quality criteria. Our proposed models achieve high correlation with human judgments, showing that it is possible to estimate summary quality  without human references. 

\section{Datasets}

We use datasets from the NIST \ducfive, \ducsix and \ducseven shared tasks \cite{DUC:2005, DUC:2006,Overetal2007}. Given a question and a cluster of newswire documents, the contestants were asked to generate a 250-word summary answering the question. \ducfive contains 1,600 summaries (50 questions x 32 systems); in \ducsix, 1,750 summaries are included (50 questions x 35 systems); and \ducseven has 1,440 summaries (45 questions x 32 systems).

The submitted summaries were manually evaluated in terms of content preservation using the Pyramid score, and according to five linguistic quality criteria ($\mathcal{Q}1, \dots, \mathcal{Q}5$), described in Figure~\ref{lqs}, that do not involve comparison with a model summary. Annotators assigned scores on a five-point scale, with 1 and 5 indicating that the summary is bad or good with respect to a specific $\mathcal{Q}$. The overall score for a contestant with respect to a specific $\mathcal{Q}$ is the average of the manual scores assigned to the summaries generated by the contestant. Note that the \ducfour shared task involved seven $\mathcal{Q}$s, but some of them were found to be highly overlapping and were grouped into five in subsequent years \cite{Overetal2007}.\footnote{The complete guidelines given to annotators for DUC 2005 and subsequent years can be found at  \url{https://duc.nist.gov/duc2005/quality-questions.txt}} We address these five criteria and use \duc data from 2005 onwards in our experiments.

\section{Methods}

\subsection{The \sumqe Model}

\begin{figure}[t]
\begin{center}
\includegraphics[width=\columnwidth]{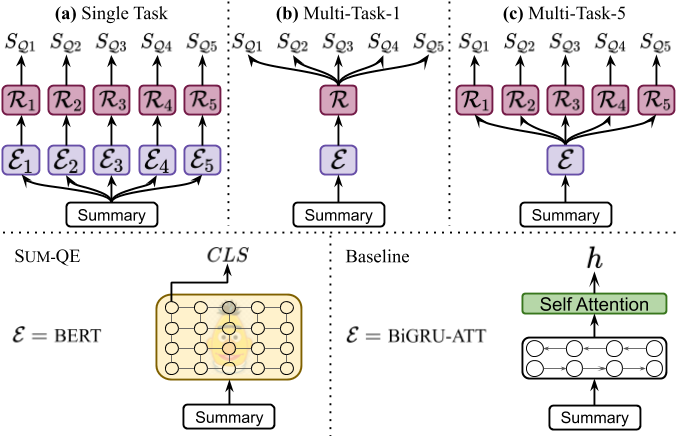}
\end{center}
\vspace*{-3mm}
\caption{Illustration of  different flavors of the investigated neural QE methods. An encoder ($\mathcal{E}$) converts the summary to a dense vector representation $h$. A regressor $\mathcal{R}_i$ predicts a quality score $S_{\mathcal{Q}i}$ using $h$. $\mathcal{E}$ is either a \bigru with attention (\bigruatt) or \bert (\sumqe). $\mathcal{R}$ has three flavors, one single-task (a) and two multi-task (b, c).}
 \label{fig:architecture}
 \vspace*{-3mm}
\end{figure}

In \sumqe, each peer summary is converted into a sequence of token embeddings, consumed by an encoder $\mathcal{E}$ to produce a (dense vector) summary representation $h$. Then, a regressor $\mathcal{R}$ predicts a quality score $S_{\mathcal{Q}}$ as an affine transformation of $h$:

\begin{equation}
    S_{\mathcal{Q}} = \mathcal{R}(h) = W^{\mathcal{R}}h + b^{\mathcal{R}} 
\end{equation}

\noindent Non-linear regression could also be used, but a linear (affine) $\mathcal{R}$ already performs well. We use BERT as our main encoder and fine-tune it in three ways, which leads to three versions of \sumqe.

\paragraph{Single-task  (BERT-FT-S-1):} The first version of \sumqe uses five separate estimators, one per quality score, each having its own encoder $\mathcal{E}_i$ (a separate \bert instance generating $h_i$) and regressor $\mathcal{R}_i$ (a separate linear regression layer on top of the corresponding BERT instance):

\begin{equation}
    S_{\mathcal{Q}i} = \mathcal{R}_i(h_i), i = 1 \dots 5
\end{equation}

\paragraph{Multi-task with one regressor (BERT-FT-M-1):} 
The second version of \sumqe uses one estimator to predict all five quality scores at once, from a single encoding $h$ of the summary, produced by a single BERT instance. The intuition is that $\mathcal{E}$ will learn to create richer representations so that  $\mathcal{R}$ (an affine transformation of $h$ with 5 outputs) will be able to predict all quality scores:

\begin{equation}
    S_{\mathcal{Q}i} = \mathcal{R}(h)[i], i = 1 \dots 5
\end{equation}
\noindent where $\mathcal{R}(h)[i]$ is the $i$-th element of the vector returned by $\mathcal{R}$.

\paragraph{Multi-task with 5 regressors (BERT-FT-M-5):} 
The third version of \sumqe is similar to BERT-FT-M-1,  but we now use five different linear (affine) regressors, one per quality score:

\begin{equation}
    S_{\mathcal{Q}i} = \mathcal{R}_i(h), i = 1 \dots 5
\end{equation}
\noindent Although BERT-FT-M-5 is mathematically equivalent to BERT-FT-M-1, in practice these two versions of \sumqe produce different results because of implementation details related to how the losses of the regressors (five or one) are combined.

\subsection{Baselines}
\label{sec:baselines}

\paragraph{\bigru{s} with attention:} This is very similar to \sumqe but now $\mathcal{E}$ is a stack of \bigru{s} with self-attention \cite{Xu2015}, instead of a BERT instance. The final summary representation ($h$) is the sum of the resulting context-aware token embeddings ($h = \sum_i a_i h_i$)  weighted by their self-attention scores ($a_i$). We again have three flavors: one single-task (BiGRU-ATT-S-1) and two multi-task (BiGRU-ATT-M-1 and BiGRU-ATT-M-5).

\paragraph{\rouge:}
This baseline is the \rouge version that performs best on each dataset, among the versions considered by \citet{Graham:2015}. Although \rouge focuses on surface similarities between peer and reference summaries, we would expect  properties like grammaticality, referential clarity and coherence to be captured to some extent by \rouge versions based on long $n$-grams or longest common subsequences. 

\paragraph{Language model (LM):} For a peer summary, a reasonable estimate of $\mathcal{Q}1$ (Grammaticality) is the perplexity returned by a pre-trained language model. We experiment with the pre-trained {GPT-2} model \cite{radford2019language}, and with the probability estimates that BERT can produce for each token when the token is treated as masked (BERT-FR-LM).\footnote{Here BERT parameters are frozen (not fine-tuned). We use the pre-trained masked \lm model to obtain probability estimates for the tokens, which are then used to calculate the perplexity.} Given that the grammaticality of a summary can be corrupted by just a few bad tokens, we compute the perplexity by considering only the $k$ worst (lowest \lm probability) tokens of the peer summary, where $k$ is a tuned hyper-parameter.\footnote{Consult the supplementary material for details.}

\paragraph{Next sentence prediction:} \bert training relies on two tasks: predicting masked tokens and next sentence prediction. The latter seems to be aligned with the definitions of $\mathcal{Q}3$ (Referential Clarity), $\mathcal{Q}4$ (Focus) and $\mathcal{Q}5$ (Structure \& Coherence). Intuitively, when a sentence follows another with high probability, it should involve clear referential expressions and preserve the focus and local coherence of the text.\footnote{We also found the three quality scores to be highly correlated. The reader may refer to the supplementary material for correlation heatmaps between the five quality scores.} We, therefore, use a pre-trained \bert model (BERT-FR-NS) to calculate the sentence-level perplexity of each summary:

\begin{equation}
    \mathcal{H}=2^{-\frac{1}{n}\sum\limits_{i=2}^n\log_2p(s_i|s_{i-1})}
\end{equation}

\noindent where $p(s_i|s_{i-1})$ is the probability that \bert assigns to the sequence of sentences $\left< s_{i-1}, s \right>$, and $n$ is the number of sentences in the peer summary.

\begin{table*}[ht!]
\centering
{
\footnotesize
\begin{tabular}{llccc|ccc|ccc}
& & \multicolumn{3}{c|}{{\bf \ducfive}} & \multicolumn{3}{c|}{{\bf \ducsix}} & \multicolumn{3}{c}{{\bf \ducseven}}
\\ 
\toprule \\[-1.4em]
& & $\rho$ & $\tau$ & $r$ & $\rho$ & $\tau$ & $r$ & $\rho$ & $\tau$ & $r$
\\ 
\toprule \\[-1.4em]
\parbox[t]{1cm}{\multirow{9}{*}{\rotatebox[origin=c]{90}{\footnotesize \parbox{2.2cm}{\centering $\mathbcal{Q}${\bf1} Grammaticality}}}}
& \rougebest     & 0.213       & 0.128  & 0.033       & -0.049       & -0.044        & 0.331       & 0.387       & 0.283       & 0.506\\
& \gpt           & 0.678       & 0.511  & 0.637       &  0.391       &  0.280        & 0.593       & 0.780       & 0.586       & 0.675\\
& \bertfrlm      & 0.437       & 0.319  & 0.025       &  0.524       &  0.354        & 0.667       & 0.598       & 0.453       & 0.566\\
\cline{2-11}\\ [-1.0em]
&  \bigruattsone  & 0.119       & 0.079  & 0.116       &  0.263       & 0.182         & 0.459       & 0.119       & 0.085       & 0.494\\
& \bigruattmone  & 0.190       & 0.144  & 0.091       &  0.619       & 0.462         & 0.757       & 0.332       & 0.235       & 0.662\\
& \bigruattmfive & 0.156       & 0.160  & 0.040       &  0.613       & 0.466         & 0.771       & 0.315       & 0.215       & 0.584\\
& \bertftsone    & 0.681       & 0.543  & {\bf 0.817} &  {\bf 0.907} & {\bf 0.760}   & {\bf 0.929} & 0.845       & 0.672       & {\bf 0.930}\\
& \bertftmone    & 0.675       & 0.543  &  0.805      &  0.889       & 0.749         & 0.902       & {\bf 0.851} & {\bf 0.684} & 0.896\\
& \bertftmfive   & {\bf 0.712} & {\bf 0.564}  & 0.802  &  0.883       & 0.732         & 0.925       & 0.840       & 0.680       & 0.902\\
\toprule \\[-1.4em]
\parbox[t]{1cm}{\multirow{7}{*}{\rotatebox[origin=c]{90}{\footnotesize \parbox{2.2cm}{\centering  $\mathbcal{Q}${\bf2} \\ Non redundancy }}}}
& \rougebest     & -0.121      & -0.081      &  0.064    & -0.401      & -0.301      & -0.408      & -0.299      & -0.222      & -0.486\\
\cline{2-11}\\ [-1.0em]
& \bigruattsone  & -0.063      & -0.049      & -0.101    &  0.511      &  0.358      &  0.514      &  0.468      &  0.352      &  0.457\\
& \bigruattmone  & -0.197      & -0.143      & -0.094    &  0.478      &  0.478      &  0.524      &  0.478      &  0.340      &  0.565\\
& \bigruattmfive & -0.226      & -0.167      & -0.124    &  0.414      &  0.304      &  0.399      &  0.283      &  0.201      &  0.238\\
& \bertftsone    &  0.330      &  0.232      &  {\bf 0.499}    &  0.677      &  0.517      &  0.679      &  0.756      &  0.576      &  0.689\\
& \bertftmone    &  0.333      &  0.232      &  0.494    & {\bf 0.791} & {\bf 0.615} & {\bf 0.789} & {\bf 0.761} & {\bf 0.596} & {\bf 0.799}\\
& \bertftmfive   & {\bf 0.377} & {\bf 0.310} & 0.471 &  0.632      &  0.460      &  0.674      &  0.754      &  0.572      &  0.740\\
\toprule \\[-1.4em]
\parbox[t]{1cm}{\multirow{8}{*}{\rotatebox[origin=c]{90}{\footnotesize \parbox{2.7cm}{\centering $\mathbcal{Q}${\bf3} \\ Referential clarity}}}}
& \rougebest     & 0.381       & 0.284       &  0.166      & 0.411       & 0.329       & 0.372       & 0.449      & 0.347        & 0.407\\
& \bertfrns      & 0.185       & 0.130       & -0.138      & 0.462       &  0.315      & 0.494       & 0.478      & 0.322        & 0.085\\
\cline{2-11}\\ [-1.0em]
& \bigruattsone  & 0.662       & 0.479       &  0.468      & 0.493       &  0.342      & 0.647       & 0.664       & 0.476       & 0.677\\
& \bigruattmone  & 0.702       & 0.540       &  0.492      & 0.527       &  0.396      & 0.681       & 0.732       & 0.533       & 0.681\\
& \bigruattmfive & 0.694       & 0.519       &  0.492      & 0.579       &  0.427      & 0.719       & 0.659       & 0.472       & 0.655\\
& \bertftsone    & {\bf 0.913} & {\bf 0.759} & {\bf 0.796} & 0.872       &  0.732      & 0.901       & {\bf 0.934} & {\bf 0.796} & {\bf 0.936}\\
& \bertftmone    & 0.889       & 0.714       &  0.761      & {\bf 0.881} & {\bf 0.735} & 0.882       & 0.879       & 0.699       & 0.891\\
& \bertftmfive   & 0.810       & 0.617       &  0.732      & 0.860       &  0.718      & {\bf 0.919} & 0.889       & 0.723       & 0.895\\ 
\toprule \\[-1.4em]
\parbox[t]{1cm}{\multirow{8}{*}{\rotatebox[origin=c]{90}{\footnotesize \parbox{2.7cm}{\centering $\mathbcal{Q}${\bf4} \\ Focus}}}}
& \rougebest     & 0.440       & 0.373       &  0.270      & 0.440       & 0.331       & 0.475       & 0.495       & 0.360       & 0.563\\
& \bertfrns      & 0.458       & 0.337       & -0.106      & 0.522       & 0.354       & 0.508       & 0.547       & 0.364       & 0.089\\
\cline{2-11}\\ [-1.0em]
& \bigruattsone  & 0.150       & 0.110       &  0.153      & 0.355       & 0.242       & 0.644       & 0.433       & 0.321       & 0.533\\
& \bigruattmone  & 0.199       & 0.118       &  0.194      & 0.366       & 0.259       & 0.653       & 0.533       & 0.372       & 0.553\\
& \bigruattmfive & 0.154       & 0.097       &  0.160      & 0.493       & 0.371       & 0.691       & 0.645       & 0.462       & 0.657\\
& \bertftsone    & 0.645       & 0.471       &  0.578      & 0.814       & 0.636       & 0.853       & 0.873       & 0.704       & 0.902\\
& \bertftmone    & 0.664       & 0.491       &  0.642      & 0.776       & 0.608       & 0.842       & {\bf 0.893} & {\bf 0.745} & {\bf 0.905}\\
& \bertftmfive   & {\bf 0.791} & {\bf 0.621} & {\bf 0.739} & {\bf 0.875} & {\bf 0.710} & {\bf 0.911} & 0.818       & 0.636       & 0.867\\
\toprule \\[-1.4em]
\parbox[t]{1cm}{\multirow{8}{*}{\rotatebox[origin=c]{90}{\footnotesize \parbox{3cm}{\centering $\mathbcal{Q}${\bf5} \\ Structure \\ \& Coherence}}}}
& \rougebest     & 0.391       & 0.300       &  0.039      & 0.080       & 0.056       & 0.023       & 0.370       & 0.292       & 0.293\\
& \bertfrns      & 0.200       & 0.153       & -0.140      & 0.171       & 0.120       & 0.285       & 0.418       & 0.280       & 0.015\\
\cline{2-11}\\ [-1.0em]
& \bigruattsone  & 0.223       & 0.153       &  0.040      & 0.458       & 0.326       & 0.526       & 0.606       & 0.442       & 0.534\\
& \bigruattmone  & 0.404 & 0.264 & 0.067 & 0.479 & 0.350 & 0.599 & 0.664 & 0.499 & 0.576\\
& \bigruattmfive & 0.244       & 0.157       & -0.113      & 0.435       & 0.296       & 0.540       & 0.522       & 0.389       & 0.506\\
& \bertftsone    & 0.536       & 0.415       &  0.477      & 0.681       & 0.522       & 0.810       & 0.862       & 0.690       & {\bf 0.857}\\
& \bertftmone    & 0.566       & 0.419       &  0.512      & 0.684       & 0.515       & 0.726       & 0.864       & 0.690       & 0.803\\
& \bertftmfive   & {\bf 0.634} & {\bf 0.472} & {\bf 0.586} & {\bf 0.796} & {\bf 0.620} & {\bf 0.892} & {\bf 0.921} & {\bf 0.787} & 0.843\\
\end{tabular}
}
\caption{Spearman's $\rho$, Kendall's $\tau$ and Pearson's $r$ correlations on \ducfive, \ducsix and \ducseven for $\mathbcal{Q}$1--$\mathbcal{Q}$5.  BEST-ROUGE refers to the version that achieved best correlations and is different across years.}
\vspace*{-4mm}
\label{tab:results}
\end{table*}

\section{Experiments}
To evaluate our methods for a particular $\mathcal{Q}$, we calculate the average of the predicted scores for the summaries of each particular contestant, and  the average of the corresponding manual scores assigned to the contestant's summaries. We measure the correlation between the two (predicted vs.\ manual) across all contestants using Spearman's $\rho$, Kendall's $\tau$ and Pearson's $r$.

We train and test the \sumqe and \bigruatt versions using a 3-fold procedure. In each fold, we train on two datasets (e.g., \ducfive, \ducsix) and test on the third (e.g., \ducseven). We follow the same procedure with the three \bigru-based models. Hyper-perameters are tuned on a held out subset from the training set of each fold.

\section{Results}

Table~\ref{tab:results} shows Spearman's $\rho$, Kendall's $\tau$ and Pearson's $r$ for all datasets and models. The three fine-tuned BERT versions clearly outperform all other methods. Multi-task versions seem to perform better than single-task ones in most cases. Especially for $\mathcal{Q}4$ and $\mathcal{Q}5$, which are highly correlated, the multi-task BERT versions achieve the best overall results. \bigruatt also benefits from multi-task learning.

\begingroup
\setlength{\tabcolsep}{2.8pt}
\begin{table}[t]
\centering
\begin{tabular}{c|c|c|c}
                     & {\bf\ducfive} & {\bf\ducsix} &  {\bf\ducseven}\\\hline
$\mathbcal{Q}${\bf1} & 3.77 ($\pm$ 0.42)      & 3.58 ($\pm$ 0.60)      & 3.54 ($\pm$ 0.78)\\
$\mathbcal{Q}${\bf2} & {\bf4.41 ($\pm$ 0.20)} & {\bf4.23 ($\pm$ 0.26)} & {\bf3.71 ($\pm$ 0.31)}\\
$\mathbcal{Q}${\bf3} & 2.99 ($\pm$ 0.50)      & 3.11 ($\pm$ 0.52)      & 3.20 ($\pm$ 0.66)\\
$\mathbcal{Q}${\bf4} & 3.15 ($\pm$ 0.41)      & 3.60 ($\pm$ 0.39)      & 3.30 ($\pm$ 0.47)\\
$\mathbcal{Q}${\bf5} & 2.18 ($\pm$ 0.46)      & 2.39 ($\pm$ 0.51)      & 2.42 ($\pm$ 0.59)\\
\end{tabular}
\caption{Mean manual scores ($\pm$ standard deviation) for each $\mathcal{Q}$ across datasets. $\mathcal{Q}2$ is the hardest to predict because it has the highest scores and the lowest standard deviation.}
\label{tab:stds}
\end{table}
\endgroup

\begin{figure}[t]
\begin{center}
\includegraphics[width=\columnwidth]{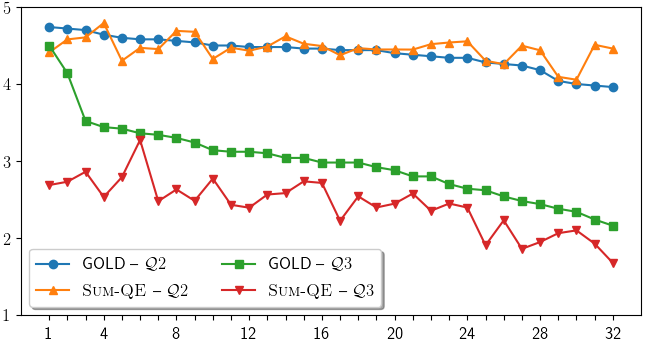}
\end{center}
\vspace*{-3mm}
\caption{Comparison of the mean gold scores assigned for $\mathcal{Q}2$ and $\mathcal{Q}3$ to each of the 32 systems in the \ducfive dataset, and the corresponding scores predicted by \sumqe. Scores range from 1 to 5. The systems are sorted in descending order according to the gold scores. \sumqe makes more accurate predictions for $\mathcal{Q}2$ than for $\mathcal{Q}3$, but struggles to put the systems in the correct order.}
 \label{fig:q2_bert_vs_gold}
 \vspace*{-3mm}
\end{figure}

The correlation of \sumqe with human judgments is high or very high \cite{hinkle2003applied} for all $\mathcal{Q}$s in all datasets, apart from  $\mathcal{Q}2$ in \ducfive where it is only moderate. Manual scores for $\mathcal{Q}2$ in \ducfive are the highest among all $\mathcal{Q}$s and years (between 4 and 5) and with the smallest standard deviation, as shown in Table~\ref{tab:stds}. Differences among systems are thus small in this respect, and although \sumqe predicts scores in this range, it struggles to put them in the correct order, as illustrated in Figure~\ref{fig:q2_bert_vs_gold}. 

BEST-ROUGE has a negative correlation with the ground-truth scores for $\mathcal{Q}$2 since it does not account for repetitions. The \bigru-based models also reach their lowest performance on $\mathcal{Q}$2 in \ducfive. A possible reason for the higher relative performance of the BERT-based models, which achieve a moderate positive correlation, is that \bigru captures long-distance relations less effectively than \bert, which utilizes Transformers \cite{Vaswani2017} and has a larger receptive field. A possible improvement would be a stacked \bigru, since the states of higher stack layers have a larger receptive field as well.\footnote{As we move up the stack, the states are affected directly by their neighbors and indirectly by the neighbors of their neighbors, and so on.}

The BERT multi-task versions perform better with highly correlated qualities like $\mathcal{Q}4$ and $\mathcal{Q}5$ (as illustrated in Figures 2 to 4 in the supplementary material). However, there is not a clear winner among them. Mathematical equivalence does not lead to deterministic results, especially when random initialization and stochastic learning algorithms are involved. An in-depth exploration of this point would involve further investigation, which will be part of future work.

\section{Conclusion and Future Work}

We propose a novel Quality Estimation model for summarization which does not require human references to estimate the quality of automatically produced summaries. \sumqe successfully predicts  qualitative aspects of summaries that recall-oriented  evaluation metrics fail to approximate. Leveraging powerful BERT representations, it achieves high correlations with human scores for most linguistic qualities rated, on three different datasets. 
Future work involves extending the \sumqe model to capture content-related aspects, either in combination with existing evaluation metrics (like Pyramid and \rouge) or, preferably, by identifying important information in the original text and modelling its preservation in the proposed summaries. This would preserve \sumqe's independence from human references, a property of central importance in real-life usage  scenarios and system development settings. 

The datasets used in our experiments come from the NIST DUC shared tasks which comprise newswire articles. We believe that  \sumqe could be easily applied to other domains. A small amount of annotated data would be needed for fine-tuning -- especially in domains with specialized vocabulary (e.g., biomedical) -- but the model could also be used out of the box. A concrete estimation of performance in this setting will be part of future work. Also, the model could serve to estimate linguistic qualities other than the ones  in the DUC dataset with mininum effort.

Finally, \sumqe could serve to assess the quality of other types of texts, not only summaries. It could thus be applied to other text generation tasks, such as natural language generation and sentence compression.
 
\section*{Acknowledgments}

We would like to thank the anonymous reviewers for their helpful feedback on this work. The work has been partly supported by the Research Center of the Athens University of Economics and Business, and by the French National Research Agency under project ANR-16-CE33-0013.

\bibliography{emnlp-ijcnlp-2019}
\bibliographystyle{acl_natbib}

\end{document}